\documentclass[sigconf,9pt,nonacm]{acmart}
\pdfoutput=1
\usepackage{amsmath,graphicx}

\usepackage{natbib}

\usepackage{xcolor}

\newcommand*\linkcolours{ForestGreen}

\usepackage{times}
\usepackage{graphicx}
\usepackage{gensymb}
\usepackage{amsmath}
\usepackage{breakurl}

\usepackage{url,hyperref}
\hypersetup{
colorlinks,
linkcolor=\linkcolours,
citecolor=\linkcolours,
filecolor=\linkcolours,
urlcolor=\linkcolours}

\usepackage{algorithm}
\usepackage{algorithmic}

\usepackage[labelfont={bf},font=small]{caption}
\usepackage{subcaption}

\usepackage{svg}

\usepackage{mathtools, cuted}

\usepackage[noadjust, nobreak]{cite}

\usepackage{tabularx}
\usepackage{amsmath}

\usepackage{tikz-cd}
\usepackage{multirow}
\usepackage{float}

\usepackage{pifont}

\newcommand*\GitHubLoc{https://github.com}

\newcolumntype{Y}{>{\centering\arraybackslash}X}

\usepackage[]{placeins}

\usepackage{placeins}

\usepackage{tikz}

\usepackage[framemethod=tikz]{mdframed}

\usepackage{afterpage}

\usepackage{stfloats}

\usepackage[normalem]{ulem}
\usepackage{atbegshi}
\newcommand{\handlethispage}{}
\newcommand{\discardpagesfromhere}{\let\handlethispage\AtBeginShipoutDiscard}
\newcommand{\keeppagesfromhere}{\let\handlethispage\relax}
\AtBeginShipout{\handlethispage}

\usepackage[colorinlistoftodos]{todonotes}

\makeatletter
\def\bstctlcite{\@ifnextchar[{\@bstctlcite}{\@bstctlcite[@auxout]}}
\def\@bstctlcite[#1]#2{\@bsphack
  \@for\@citeb:=#2\do{%
    \edef\@citeb{\expandafter\@firstofone\@citeb}%
    \if@filesw\immediate\write\csname #1\endcsname{\string\citation{\@citeb}}\fi}%
  \@esphack}
\makeatother

\usepackage{comment}
\usepackage{booktabs,chemformula}

\title{Flexible Multiple-Objective Reinforcement Learning \\ for Chip Placement}
\author{\normalsize Fu-Chieh Chang$^{1,3}$, Yu-Wei Tseng$^2$, Ya-Wen Yu$^2$, Ssu-Rui Lee$^2$, Alexandru Cioba$^1$, I-Lun Tseng$^2$, Da-shan Shiu$^1$, Jhih-Wei Hsu$^3$, Cheng-Yuan Wang$^3$, Chien-Yi Yang$^{3}$, Ren-Chu Wang$^{3}$, Yao-Wen Chang$^3$, Tai-Chen Chen$^4$ and Tung-Chieh Chen$^4$}

\affiliation{ \footnotesize{$^1$MediaTek Research, $^2$MediaTek Inc., $^3$National Taiwan University, $^4$Maxeda Technology Inc.}
 }
 
\affiliation{%
  \footnotesize $^1$\{mark-fc.chang, alexandru.cioba, ds.shiu\}@mtkresearch.com , 
  $^2$\{yu-wei.tseng, yw.yu, ssr.lee, i-lun.tseng\}@mediatek.com
  $^3$\{r09943095, r00946007, b06901040, b06901038, ywchang\}@ntu.edu.tw
  $^4$\{taichen, tungchieh\}@maxeda.tech 
}

\acmConference[DAC '22]{Design Automation Conference 2022}{July 10--14, 2022}{San Francisco, CA}

\begin{document}

\bstctlcite{IEEEexample:BSTcontrol}

\captionsetup{skip=0pt}

%

%

\renewcommand{\baselinestretch}{0.96}

\begin{abstract}

Recently, successful applications of reinforcement learning to chip
placement have emerged. Pretrained models are necessary to improve
efficiency and effectiveness. Currently, the weights of objective
metrics (e.g., wirelength, congestion, and timing) are fixed
during pretraining. However, fixed-weighed models cannot generate
the diversity of placements required for engineers to accommodate
changing requirements as they arise. This paper proposes flexible
multiple-objective reinforcement learning (MORL) to support objective
functions with inference-time variable weights using just a single
pretrained model. Our macro placement results show that MORL
can generate the Pareto frontier of multiple objectives effectively.
\end{abstract}


\maketitle

\section{Introduction}\label{intro}


With the evolution of AI, 5G, and high-performance computing, 
there is a growing demand for chips that can quickly process huge amounts of data, 
and such chips require a lot of memory or SRAM macros for the temporary storage of data. 
Additionally, with the advance of semiconductor manufacturing technologies,
the numbers of SRAM macros and standard cell instances in a chip keep increasing.
Nowadays, an advanced SoC design can contain more than one thousand macros 
and millions of standard cell instances.


Completing chip designs manually is not only time-consuming, but also virtually impossible for the complexity of modern SoC designs.
Therefore, design automation tools are required to help IC design engineers cope with the challenges.
Furthermore, in the physical design phase of a chip design project, the chip placement process (which is the process of determining the locations of macros and standard cell instances within a certain circuit block) is one of the most critical processes affecting the PPA (power, performance, and area) or even the final specification of a chip design.


In recent years, reinforcement learning (RL) has been proved to be a ubiquitous tool for the optimization of complex, non-differentiable objectives, subject to noise or uncertainty in the signals. Typically, these algorithms operate in a closed-loop fashion, with many iterations of signal feedback being required to improve results. This problem is exacerbated in environments with complex transition dynamics and noisy signals. Deep RL (DRL) has presented a long list of success stories in addressing this problem, with the relatively inexpensive optimization of stochastic gradient descent and complex policy parametrization leading the fore in solving previously intractable decision making problems \citep{SilverHuangEtAl16nature}. However, even state-of-the-art DRL methods suffer from many drawbacks. Neural network generalization is a critical component of good performance, but only recently has there been a success in this direction \citep{bartlett2021deep}. Furthermore, the problem of generalizing across environments with different transition dynamics and reward signals in RL is even less well studied \citep{vanseijen2017hybrid, dennis2021emergent}. DRL optimization is often brittle and requires complex hyperparameter tuning \citep{khadka2018evolutionguided}, and these issues are only made worse by the long training times, poor sample efficiency and expert supervision required.   
As a typical solution to some of these issues, pretrained representations of states, actions, or the world dynamics aid in reducing runtime during execution, where only a variable amount of fine-tuning takes place. 

For the macro placement problem, these issues translate into a barrier to the adoption of RL techniques. In a keynote speech at ISSCC 2020 ~\citep{DBLP:conf/isscc/Dean20}, the importance ``[of allowing] ASIC designers to quickly generate many alternatives with different tradeoffs of area, timing, etc.'' is emphasized. 
However, such flexibility is not available for models trained to optimize a single objective. Fixed-weight models cannot generate the diversity of placements required for engineers to accommodate changing PPA requirements as they arise. Furthermore, the adoption of RL in placement is additionally hampered by the typical runtime of a QoR computation (wirelength and routing congestion) for traditional place-and-route tools. A solution is using weaker proxy estimates, but this adversely impacts the representation learning in DRL. Therefore, high-speed placement and efficient congestion analysis methods are required.


Our contributions in this paper are: 
(1) A flexible multiple-objective reinforcement learning (MORL) to support objective functions with inference-time variable weights using a single pretrained model. 
In the reward function, 
we have considered objectives of wirelength, congestion, and user anchors. 
(2) A cluster size selection method that achieves a high correlation to the original unclustered netlist is proposed.
Performing placement and congestion analysis on a clustered netlist is fast and maintains high-fidelity reward calculations. (3) An adaptation of proximal policy gradients to the MORL setting. 

This paper is organized as follows. Previous works on mixed-size placement are introduced in Section \ref{lit-rev}.
Mixed-size placement modeled as an RL problem and background of multi-objective reinforcement learning are described in Section ~\ref{background}.
The proposed MORL algorithm for mixed-size placement is introduced in Section~\ref{algo}.
Experimental results presented in Section~\ref{experiments} demonstrate that MORL can generate the Pareto frontier of multiple objectives effectively. Furthermore, we recover human-level performance when users' preferences match previously known EDA objectives.

\section{Related work}\label{lit-rev}

Mixed-size placement refers to placing modules (macros and standard cell instances) onto a chip canvas.
Modern designs often contain large numbers of macros and huge numbers of standard cell instances. 
Because macros are typically orders of magnitude larger than standard cells, handling of non-overlapping constraints among the modules presents a unique challenge.
Works on mixed-size placement fall into three categories: analytical methods, packing-based methods, and machine-learning-based methods. 

Analytical methods~\citep{DBLP:journals/tcad/ChengKKW19,DBLP:journals/tcad/LinJGLDRKP21,DBLP:conf/dac/LinLW19,DBLP:conf/iccad/ChenCC17} model the wirelength and congestion in the objective function under module overlap constraints. 
They model the non-uniformity of the module distribution as a penalty term to solve a constrained minimization problem mathematically. 
The overlap penalty is gradually increased so that the minimizer can spread modules into the placement area to obtain close-to-overlap-free mixed-size placement results. 
To obtain a legal placement result, an additional legalization step is required for overlap removal.

Packing-based methods~\citep{DBLP:journals/tvlsi/LinDYCL21,DBLP:conf/aspdac/LiuCCK19} search the macro overlap-free solution space using combinatorial methods. 
Simulated annealing (SA) is one of the most popular approaches in this category. 
Although SA has the advantage of incorporating various placement objectives, it has scalability issues and thus cannot handle large designs effectively. 
Also, it requires an initial macro and cell placement as the reference placement positions. Since only macros are considered during packing,
the packing-based method cannot estimate cell-related objectives (e.g. wirelength and congestion) accurately. 
The objective function of a packing-based method often includes certain human expert knowledge; for example, macros can be forcefully placed close to either the corner or the chip boundary, and adding irregularity penalty into objective terms to obtain array-like macro placements. This requires fine-tuning of the objective terms to obtain high-quality placement results.

A state-of-the-art placement tool that employs the power of machine learning, deep reinforcement learning (DRL), and representation learning has been developed by Mirhoseini et. al \citep{goog-rl-eda-nature,mirhoseini2020chip}. 
In this work, 
the authors deploy a fully AI-driven solution for the placement problem, 
without the chip placement knowledge from human experts. 
The algorithm pretrains a placement policy from many ICs,
leverag ing graph structures of various IC netlists through a Graph Neural Network (GNN). 

Multiple-objective RL also has a long history and applications in a wide range of fields. Deep RL techniques, however, have only been relevant in this direction recently, and furthermore, due to the function approximation properties of neural networks, have mainly been studied in the context of \textit{off-policy} RL. In \citep{yang2019generalized}, a DQN approach for MORL computes multi-objective $Q$-value function approximations. In \citep{2020MODRL}, the authors propose a general framework for all off-policy DRL optimization strategies. In \citep{vanseijen2017hybrid}, DQNs are again revisited with a focus on generalization across domains. MORL is cast as a mixture of expert synthesis problems with behaviour cloning in \citep{abdolmaleki2021multiobjective}.

\section{Problem setting}
\label{background}

\subsection{Mixed-size placement}\label{notation}
In the field of the VLSI physical design automation, a placement problem refers to placing a set of nodes onto a variable-size chip canvas so as to minimize a variety of placement costs, such as the total wirelength, routing congestion, timing violations, and power. A node can refer to either a macro (either a pre-placed macro or a macro that needs to be placed) or a standard cell instance/cluster. In the current definition of the mixed-size placement problem, only macros and standard cell instances/clusters need to be placed.
The real EDA placement problem is abstracted into a synthesized netlist (e.g., Verilog file), a physical layout library (e.g., LEF file), and a chip canvas (e.g., DEF file).
These files contain the information of node features (such as node types and their physical dimensions), net/wire connections, features of the canvas, and locations of the I/O ports. Typically, after performing placement, an EDA tool will be able to compute the final placement metrics, which can include the total wirelength, routing congestion, timing, and power.

In \citep{goog-rl-eda-nature}, the aforementioned placement workflow is modeled as a reinforcement learning (RL) optimization problem, where the macros are placed by the policies guided by an EDA reward signal. Because a typical EDA tool can offer trade-offs between speed and accuracy in terms of placement metric computations, and the reward signal represents a critical guide for the experience collection process in RL, \citep{goog-rl-eda-nature} employs a proxy computation of these metrics (see Section~\ref{metrics}). 
The reinforcement learning policy is thus trained to place macros only, and the cell placement is carried out by a classical (non-AI) tool such as a force-directed placement tool.



\begin{figure}[h]
\centering
\includegraphics[width=0.65\columnwidth]{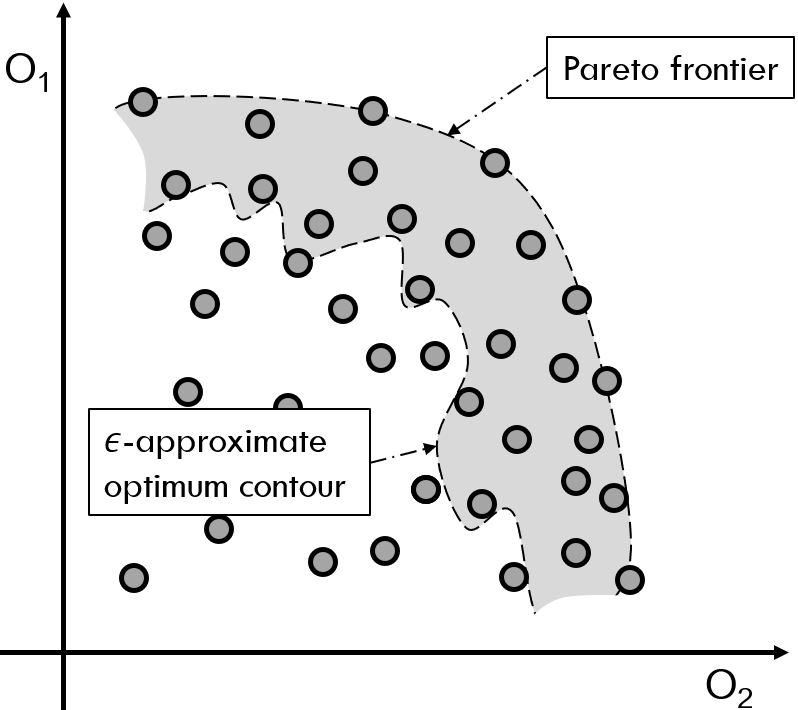}
\caption{Pareto frontier for a 2-dimensional objective simplex corresponding to $O_1$ and $O_2$ and $\epsilon$ - approximate contour. Transitioning from the approximation to the actual optimum requires only a fraction of the computational time.
}
\label{fig:pareto-abstract}
\end{figure}
\subsection{Multi-objective reinforcement learning}\label{morl-notation}
In what follows we will refer to an MDP as a tuple $(\mathcal S, \mathcal A, p, r)$ consisting of state and action space, transition dynamics $p$, and associated reward signal $r$. The standard description can be found in \citep{Sutton1998}.
We define a \textit{multi-objective MDP} (MOMDP), $\mathcal M$,  to be an MDP with state space $\mathcal S$ and action space $\mathcal A$ and fixed transition dynamics, but with a set of reward signals indexed by $i$: reward signal $i$ is denoted as $\mathbf r = r^{(i)}$. For such a vector of reward signals, a preference parameter, $\mathbf{\omega} = (\omega_i)_{i=1}^K$, can be used to collapse the MOMDP $\mathcal M$ to a standard MDP $\mathcal M^\omega$, by recovering a single reward signal $r_\omega = R(\mathbf r, \mathbf{\omega})$ for some function $R$.  Typically, and in the rest of this text, a convex linear combination is used $r_\omega = \sum_{i=1}^K r^{(i)} \omega_i$, where $\omega \in (\Omega^K, \mu)$, the standard $K$-dimensional simplex equipped with a probability measure $\mu$, typically the uniform measure. 
States and actions in $\mathcal M^\omega$ will be indexed  by $\omega$ as in $(s;\omega)$ and $(a;\omega)$ respectively. 
We adopt the setting of episodic RL, where there is a well-defined initial state, $s_0$ which is independent of $\omega$.

Solving an MOMDP amounts to solving each individual MDP $\mathcal M^\omega$. Such as solution amounts to finding a continuum of policies $\pi_\omega$ which maximizes the value of the initial state, $J(\omega, \pi_\omega) = V_{\pi_\omega}(s_0; \omega)$ for each value of $\omega$.
To make the problem tractable, define an \textit{parametric policy family} to be a set of policies $\pi_{\omega,\theta}$ continuously parametrized by $\theta$ .  

For each $\mathcal M^\omega$, within each parametric policy family, there is an optimal policy $\pi^*_\omega$ corresponding to a parameter $\theta^*(\omega)$, which is not necessarily unique. 
If we can find $\theta^*$ as a solution to $\max_\theta J(\theta, \omega)$ where $J(\theta, \omega)$ is defined as
$J(\theta, \omega) := V_{\pi_{\theta, \omega}}(s_0; \omega)$,
this represents a \textit{Pareto optimal choice} of $\theta$ within the parametric family. Equivalently, $\max_\theta J(\theta, \omega)$ admits a solution $\theta^*(\omega) = \theta^*$, constant in $\omega$. In practice, we will relax the requirement of finding solutions to an approximation. 
An $\epsilon$\textit{-approximate Pareto optimum contour} within a parametric family $\pi_{\omega,\theta}$ is a choice of $\theta^*$ which, for all $\omega$ satisfies
\begin{equation}
    |J(\theta^*, \omega) - \max_\theta J(\theta, \omega)| < \epsilon
\end{equation}
Figure \ref{fig:pareto-abstract} showcases a qualitative representation of the difference between the Pareto frontier and an $\epsilon$-approximate optimum. At an $\epsilon$-approximate optimal configuration, all reward objectives may still be improved simultaneously.

\section{Methodology}\label{algo}

\subsection{Placement objectives and approximators}\label{metrics}


We consider the most important placement objectives, wirelength, congestion, and timing, in our reward approximators. Since a reward function is evaluated many times during RL training, we need to have metrics that can be quickly evaluated while keeping good correlations to the final placement objective functions.



The most widely used wirelength model is the total half perimeter wirelength (HPWL). The wirelength reward $\ell_{WL}$ is the summation of all nets $n$ using their half perimeter length of the minimum bounding box that can enclose all pins associated with net $n$, as computed in the following manner, 
\begin{equation}
\ell_{WL} = -\sum_{n}{ ( \max_{pin_i,pin_j \in n} | x_i - x_j | + \max_{
pin_i,pin_j \in n} | y_i -y_j | ) },
\label{equ:wirelength}
\end{equation}
where $(x_i,y_i)$ is the coordinate of $pin_i$.
In addition to wirelength, a routing congestion metric is used to ensure that a placement result can indeed be routed.
We adopt a similar congestion metric from DAC 2012 routability-driven placement contest~\citep{DBLP:conf/dac/ViswanathanASLW12}. Let $ACE$ be the average congestion g-cell edges based on the histogram of g-edge congestion as described in~\citep{DBLP:conf/dac/ViswanathanASLW12,Wei:DAC2012}.  Then, $ACE(k)$ computes the average congestion of the top $k$\% congested g-cell edges. We calculate the peak-weighted congestion ($\ell_{C}$) as \\
\begin{equation}
\label{equ:congestion}
\ell_{C} = -\frac{ACE(k)}{4}, k \in 0.5, 1, 2, 5.
\end{equation}
We apply the same parameters (0.5\%, 1\%, 2\%, and 5\%) used in the previous works because this setting is proven to be effective on many benchmark cases.
We first apply a global routing, and then compute $\ell_{C}$ as our congestion metric. Note that the g-cell size is adjusted based on the average cell cluster dimensions with pin locations approximated as the centers of cell clusters.
In modern VLSI placement, additional reference anchors need to be supported to control the final placement to meet layout implementation requirements, such as fence regions~\citep{DBLP:conf/ispd/BustanyCSY15} and datapath alignment constraints~\citep{DBLP:conf/dac/HuangLLCWY17}. Therefore, an additional anchor-distance objective is enforced, with the aim of meeting specific implementation-level guidelines.
We enforce an anchor displacement:
\begin{equation}\label{anchor-dist}
    \ell_{A}=-\sum_{i}a_i\sqrt{(X_i-x_i)^2+(Y_i-y_i)^2},
\end{equation}
where $(X_i, Y_i)$ is the user-defined target coordinate, $(x_i, y_i)$ is the coordinate of the center of each macro, and $a_i$ is the weight of each anchor.
We define the EDA objective as:
\begin{equation}\label{EDA-obj}
    \ell_{EDA} = \alpha \ell_{WL} + \beta \ell_C + \delta \ell_A, 
\end{equation}
where $\alpha$, $\beta$, and $\delta$ are weights to control the significance of each corresponding objective; the sum of the first two terms is denoted by $\ell_{WL+C}$.
For modern chip designs, engineers may need to explore different placement styles in the solution space.

\subsection{Solving MOMDPs with policy gradients}

How should one attempt to find $\epsilon$-approximate Pareto optimal contours? The objective $\hat J(\theta) = \mathbb E_\omega [J(\theta;\omega)]$
when optimized, will produce the best average contour, but cannot guarantee $\epsilon$-optimality. The structure of the policy $\pi_{\theta, \omega}$ and in particular its dependence on the parameters is what determines whether the reward landscape $J(\theta;\omega)$ admits a global $\epsilon$-approximate optimal contour.
We will iteratively approximate the maximum of $\hat J(\theta)$ by
$\theta_{n+1} = \theta_n + \alpha_n \nabla_{\theta_n} \hat J(\theta)$.
The gradient $\nabla_{\theta} \hat J(\theta)$ can be obtained through standard policy gradient estimates, if $J$ is uniformly bounded and regular enough (has Lipschitz gradients).
In our application, however, we will be employing the PPO gradient estimator with generalized advantage estimation (GAE), see \citep{schulman2015high, schulman2017proximal}. We choose PPO as a basis of our algorithm following its success in the placement problem in \citep{goog-rl-eda-nature}.
In the MORL setting, the resulting procedure will output a parametrized policy family for all preferences, optimized on-policy, through multi-objective proximal policy optimization (MOPPO).

Adaptation of PPO to MORL, however, is challenging both in terms of setting the correct neural architectures as well as adapting the training regime. 
The standard PPO loss encompasses two other components, namely a value loss which controls the error in value function estimates as well as an entropy loss which is used to constrain the uncertainty in the distribution $\pi_{\theta, \omega}(s)$. 
The value function $v(s; \theta_v)$ is modified to receive the parameter $\omega$ as input (i.e. $v(s, \omega; \theta_v)$) and the value loss will be computed across input states and values of $\omega$ sampled from the buffer. This is similar to the $Q$-function update in \citep{yang2019generalized}.
The entropy loss is already an average of entropy values of the policy head across states, so we  further average over the second component of the state for each $\mathcal M^\omega$, the parameter $\omega$.
The value function $\mathbf{V}^{\pi}$ outputs a $K$-dimensional vector, one corresponding to each of the reward objectives:
\begin{equation}
\mathbf{V}^{\pi_{\theta, \omega}}(s,\omega)= \mathbb{E}[\sum_{k=0}^{\infty}\gamma^k \mathbf{r}(s_{t+k},\pi_{\theta, \omega}(s_{t+k})) |s_t=s]
\end{equation}
where $\gamma$ is a discounting factor. The estimated GAE advantage of a given length-$T$ trajectory can be written as (cf. \citep{schulman2015high}): 
\begin{equation}
\hat{\mathbf{A}}_t(\omega) =\delta + (\gamma\lambda)\delta_{t+1}+ \cdots +(\gamma\lambda)^{T-t+1}\delta_{T-1}\\
\end{equation}
\noindent where $\delta_t = \mathbf{r}_t+\gamma\hat{\mathbf{V}}(s_{t+1},\omega) - \hat{\mathbf{V}}(s_t,\omega)$
and $\hat{\mathbf{V}}(s_t,\omega)$ is a running estimate for the value function. Recall the importance sampling ratio $\rho_t(\theta,\omega)=\pi_{\theta_{current},\omega}(a_t|s_t)/\pi_{\theta_{old},\omega}(a_t|s_t)$.
Then the PPO policy loss component becomes:
\begin{align}
L^{CLIP}(\theta,\omega) = \hat{\mathbb{E}}_t \Big[\min(\rho_t(\theta,\omega)\omega^T\hat{\mathbf{A}}_t(\omega),\\,
\text{clip}(\rho_t(\theta,\omega),1-\epsilon,1+\epsilon)\omega^T\hat{\mathbf{A}}_t(\omega)) \Big]
\end{align}
\noindent where $\epsilon$ is the importance sampling threshold. The entire loss includes the value loss and entropy terms:
\begin{align}\label{loss}
L(\theta,\omega) = \hat{\mathbb{E}}_t\Big[L_t^{CLIP}(\theta,\omega)+c_1L_t^{VF}(\theta,\omega)-c_2S[\pi_{\theta,\omega}](s_t,\omega)\Big]
\end{align}
where 
\begin{equation}
    L_t^{VF}=|\omega^T \mathbf{V}^{\pi_{\theta, \omega}}(s_t,\omega)- \omega^T \hat{\mathbf{V}}(s_t,\omega)|
\end{equation}

\noindent and $S[\pi_{\theta,\omega}]$ is the entropy of the policy, and $c_1$, $c_2$ are hyperparameters which can be tuned to aid optimization. 

\subsection{Implementation and neural architectures}

In \citep{goog-rl-eda-nature}, architecture implementation follows the following schema:

\begin{tikzcd}
& & \text{Action space} \\
\text{Input data} \ar[r,"GNN", "MLP"'] & \text{Hidden state} \ar[ru,"Policy"] \ar[rd,"Value Function"] &  \\
& & \text{Value}
\end{tikzcd}

The input data, including the circuit netlist, index of macro to be placed, canvas metadata is processed through a message-passing graph neural network as well as several embedding modules and feed-forward networks to encode the data into a single 32 dimensional hidden state representation. The policy is a deconvolutional CNN which maps the hidden state into a discrete space, corresponding to a discretization of canvas positions. The value function is a feed-forward net. 
We modify this architecture in two ways. First, the preference $\omega$ is mapped through a non-linearity an concatenated into the intermediate layers of the state-representation. The hidden state maintains its dimension of 32. Secondly, the value-function is copied $K$ times to make up the vector-valued value function $\hat{\mathbf{V}}(s,\omega)$, but weights are allowed to optimize independently.

\section{Experiments}
\label{experiments}



\subsection{MOPPO training}
As a modification of an online RL algorithm, optimization of Eq.~\ref{loss}, takes place in alternating phases of experience collection and parameter updates. During experience collection, the preference parameter is sampled $\omega \sim (\Omega,\mu)$. Episodes are thus collected independently, and the experience buffer is populated with data from all individual episodes. Rewards are attached to the final state in episode collection, as described in Section~\ref{reward_comp}, rewards for intermediate states being 0. Advantages are computed and then data is randomly shuffled for batching and policy updates. Episode lengths for Superblue designs (see Section~\ref{superblue}) vary from 84 to 294, which makes the reward signal highly sparse. Experience buffers are populated with 1,000 episodes each. Policy updates take place with a batch size of 512 distributed across 8 GPUs over 6 epochs. Hyperparameters of training are the same as in \citep{goog-rl-eda-nature}, with two minor additions to stabilize training, namely the addition of a scheduler 
for the learning rate (on top of the adaptive learning rates of the Adam optimizer).

Most DRL applications where cross-environment generalization is required, employ pretraining. This is the case in \citep{goog-rl-eda-nature} where pretrianing is followed by fine-tuning. Preraining is conducted differently in our case, namely, instead of pretraining across multiple designs, with the aim of cross-canvas generalization, our aim is for $\pi_{\theta^*, \omega}$ to be approximately optimal on all values of $\omega$, namely to produce approximately optimal placements in all states $(s, \omega) \in \mathcal S_{\mathcal M^\omega}$. Therefore, at this stage we maintain the MDP transition dynamics constant, i.e. we employ a single canvas. Instead of fine-tuning, to demonstrate the generalization properties of our model, we present zero-shot inference results as in \citep{dennis2021emergent}. For the preference space $\Omega$, following \citep{goog-rl-eda-nature}, we fix parameters $(\alpha, \beta) = (0.991, 0.009)$ in Eq.~\ref{EDA-obj} and construct the preference space and preference parameters $\omega$=$(\omega_1,\omega_2)$ as 
\begin{equation}
    \Omega = \{\omega_1(\alpha \ell_{WL} + \beta \ell_C) + \omega_2  \ell_A | \omega_1,\omega_2 \in [0,1], \omega_1+\omega_2=1\}.  
\end{equation}
We uniformly sample episodes of experience distributed across 200 CPU threads and train until a relative reward improvement threshold is met, in practice this takes between 300 and 400 thousand episodes. During training, we preferentially resample $\omega$'s with low returns to improve the average returns across a buffer.

To demonstrate the effect of the anchor-distance objective $\ell_A$ on the macro-placement result, we fix an anchor lying in the center of the canvas (i.e., $(X_i, Y_i) := (W/2, H/2)$ where $W, H$ are the canvas width and height respectively). The anchor weight $a_i$ is set to the square root of the area of each macro. In addition, to normalize the magnitudes of $\ell_{WL}$, $\ell_C$ and $\ell_A$, we collect 1000  placements from an initialized network, and then average the values of $\ell_{WL}$, $\ell_C$ and $\ell_A$ from these 1000 placements as the normalization constants of these three objectives respectively. 


\subsection{Dataset} \label{superblue}

We test MOPPO using the ICCAD 2015 Superblue benchmark suite~\citep{DBLP:conf/iccad/KimHLV15}, which consists of 8 circuits. The number of macros ranges froms 84 to 294, and the number of standard cells ranges from 770k to 1930k. We focus on Superblue10 and Superblue18, which are the two designs that contain only rectangular macros. Since our method is implemented to handle rectangular macros primarily, these two cases provide the only fair comparison to tapeout results.
On other designs, results remain qualitatively similar.
We plan to adapt our algorithm to correctly place non-rectangular macros in future work.


\begin{figure}[htbp]
  \includegraphics[width=8cm]{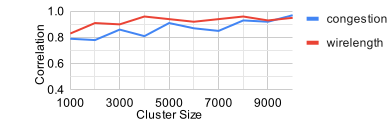}
  \caption{Correlation between clustered and unclustered wirelength/congestion (red/blue lines) vs.~number of clusters for Superblue10. Correlation increases with number of clusters; smaller cluster sizes induce faster reward computation.}
  \label{fig:cluster_size}
\end{figure}

\begin{figure*}[h]
     \centering
     \begin{subfigure}[b]{0.12\textwidth}
         \centering
         \includegraphics[width=\textwidth]{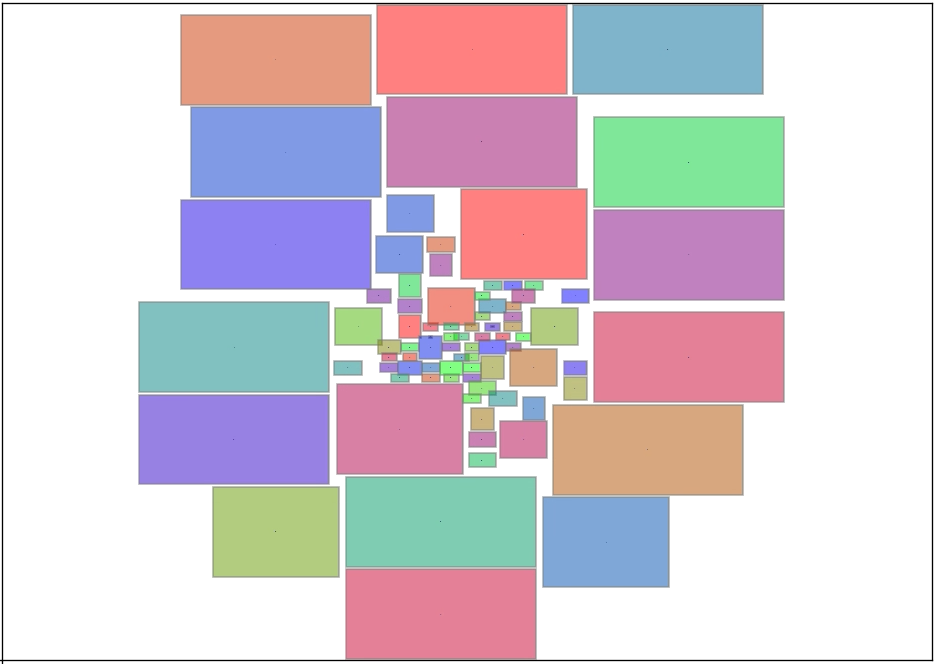}
        \includegraphics[width=\textwidth]{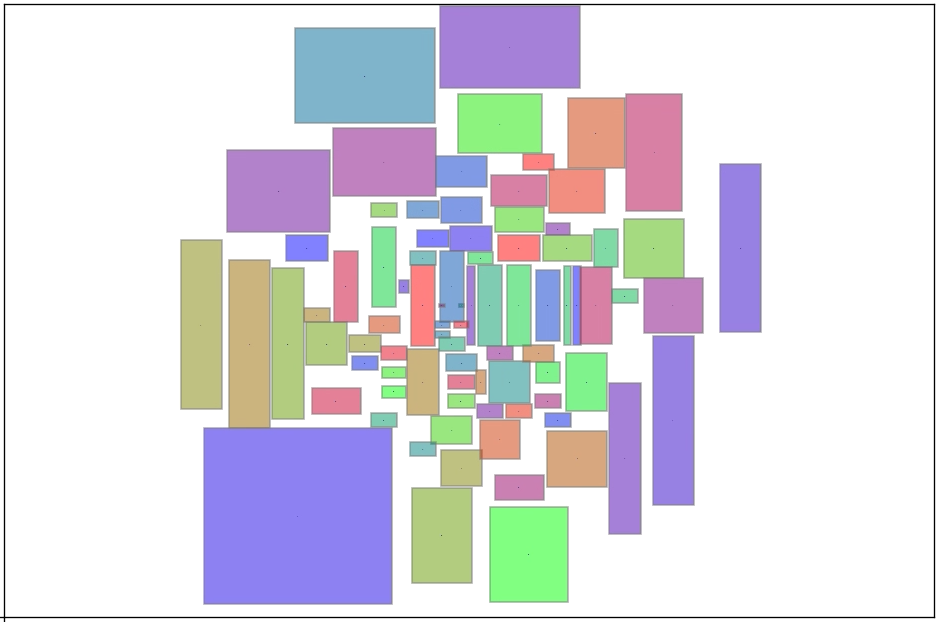}
         \caption{fixed preference \\ $\omega = (0,1)$.}
     \end{subfigure}\hfill
     \begin{subfigure}[b]{0.12\textwidth}
         \centering
         \includegraphics[width=\textwidth]{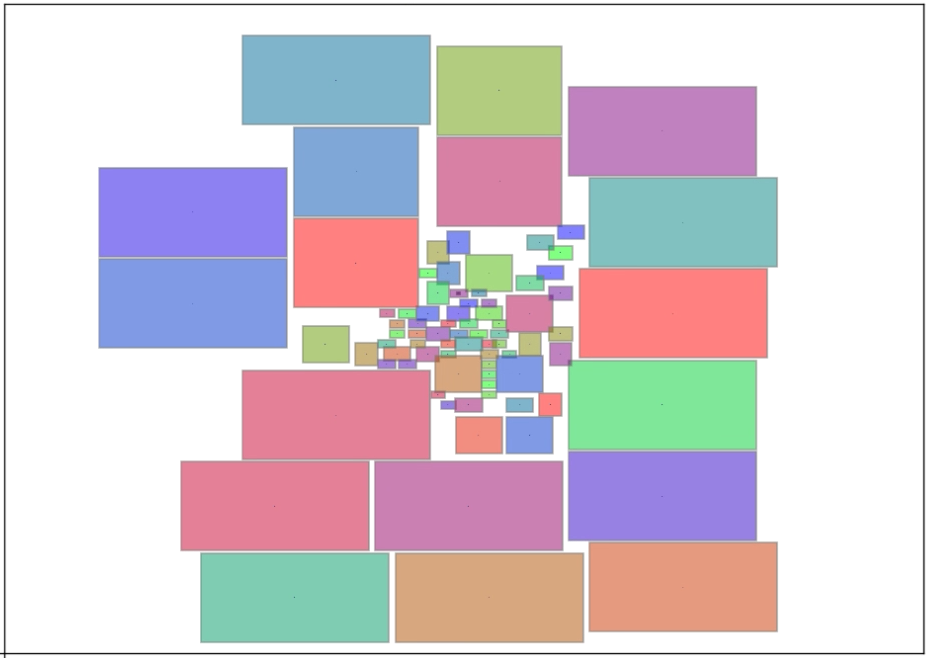}
        \includegraphics[width=\textwidth]{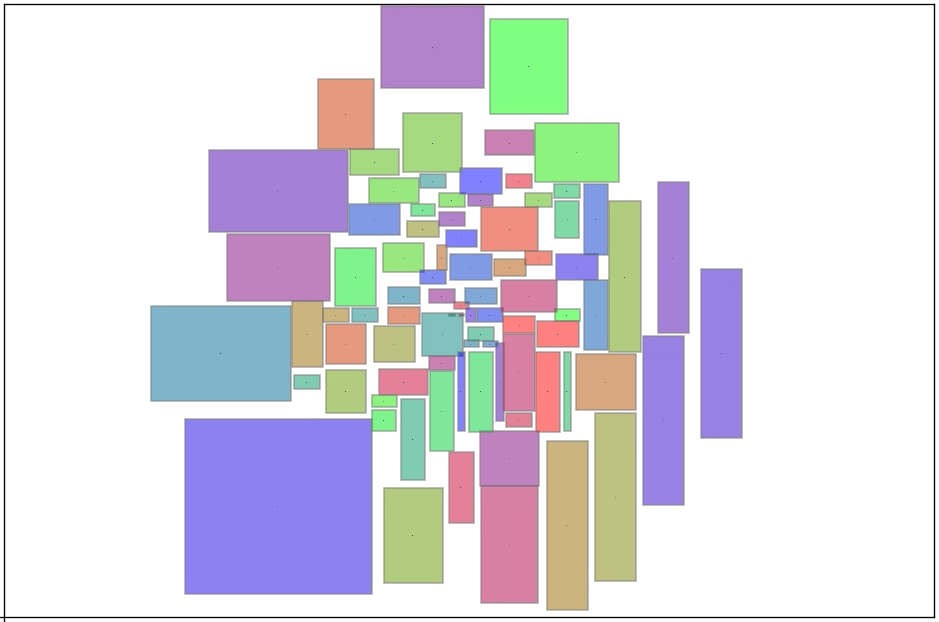}
         \caption{MOPPO \\ $\omega=(0,1)$.}
     \end{subfigure}\hfill
     \begin{subfigure}[b]{0.12\textwidth}
         \centering
        \includegraphics[width=\textwidth]{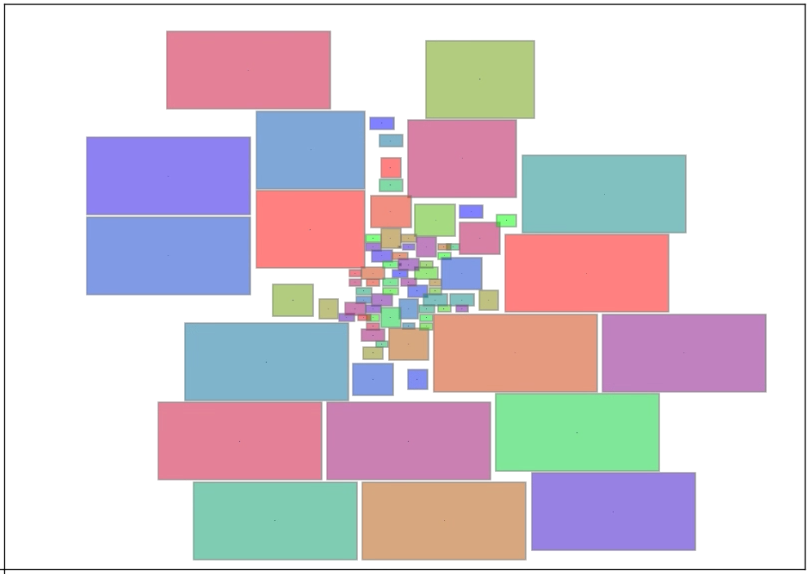}
        \includegraphics[width=\textwidth]{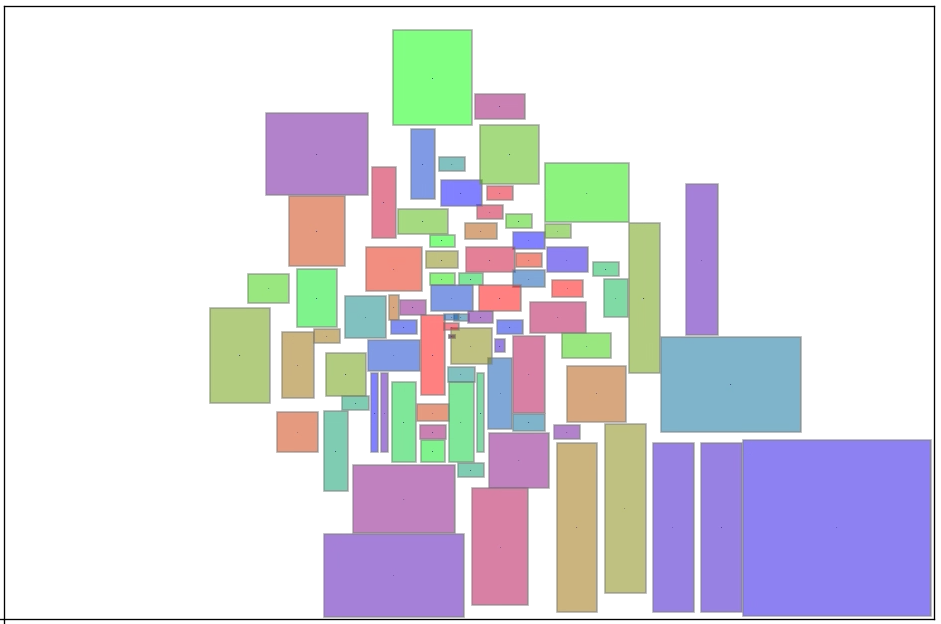}
         \caption{MOPPO\\ $\omega=(0.3,0.7)$.}
     \end{subfigure}\hfill
     \begin{subfigure}[b]{0.12\textwidth}
         \centering
         \includegraphics[width=\textwidth]{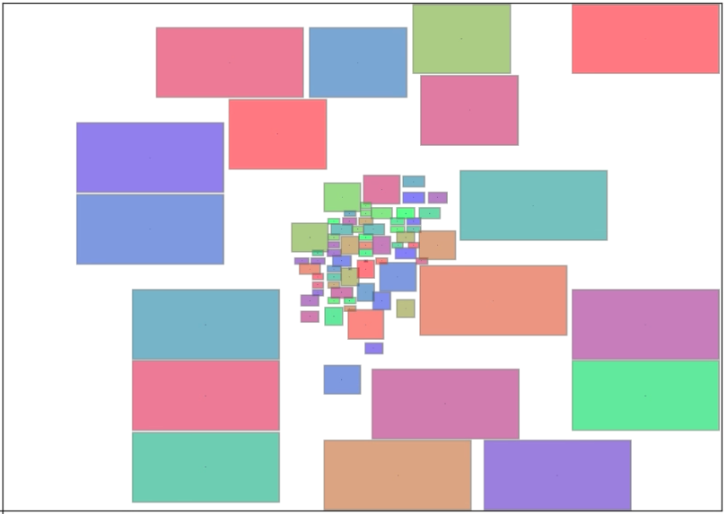}
         \includegraphics[width=\textwidth]{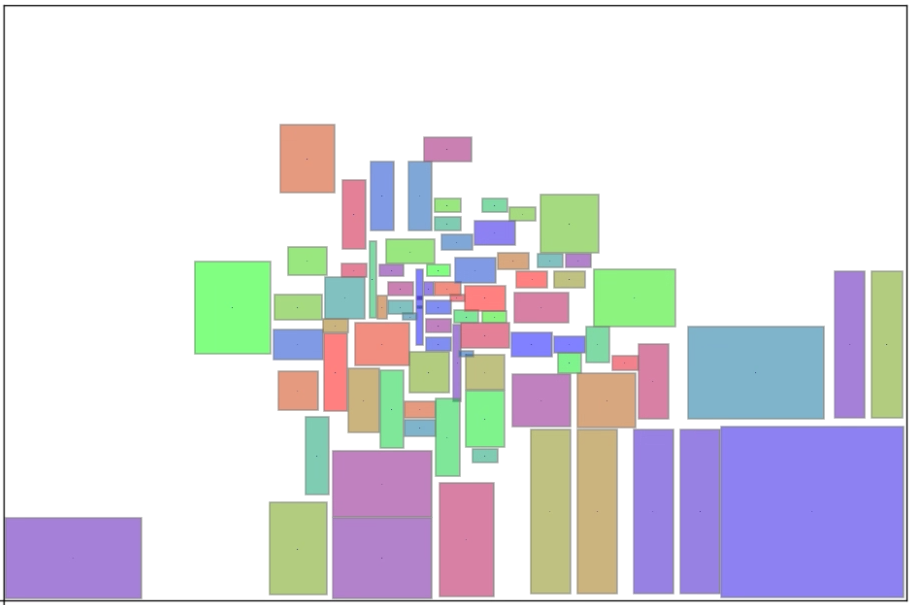}
         \caption{MOPPO \\ $\omega=(0.5,0.5)$.}
     \end{subfigure}\hfill
     \begin{subfigure}[b]{0.12\textwidth}
         \centering
         \includegraphics[width=\textwidth]{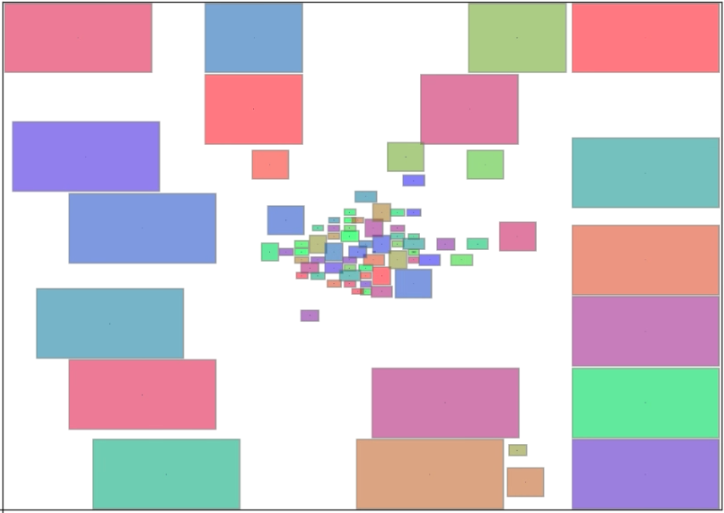}
         \includegraphics[width=\textwidth]{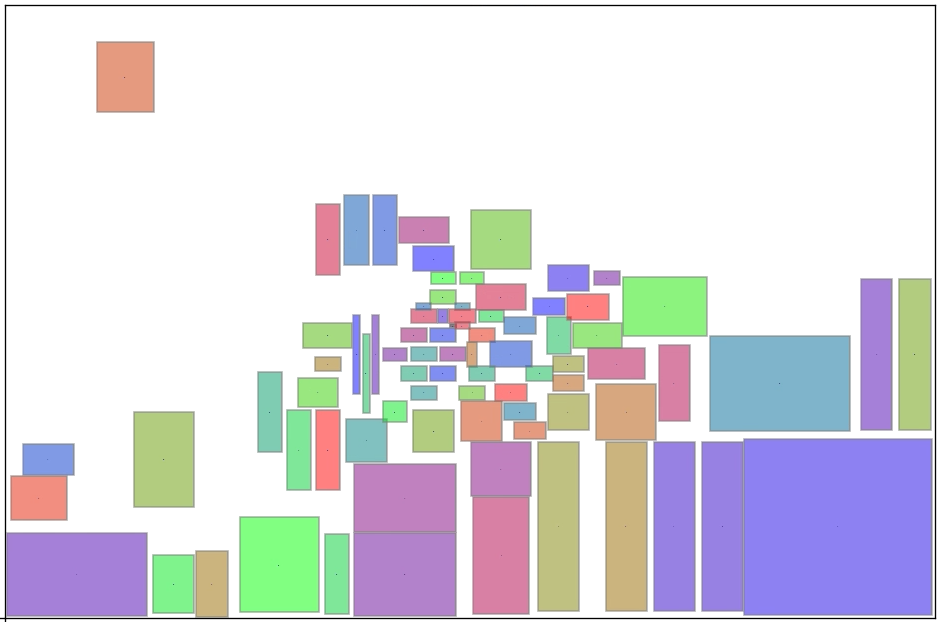}
         \caption{MOPPO \\ $\omega=(0.7,0.3)$.}
     \end{subfigure}\hfill
     \begin{subfigure}[b]{0.12\textwidth}
         \centering
         \includegraphics[width=\textwidth]{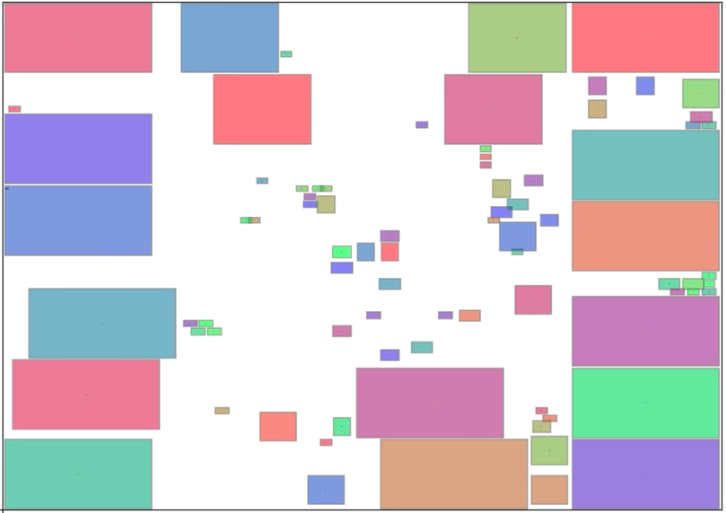}
         \includegraphics[width=\textwidth]{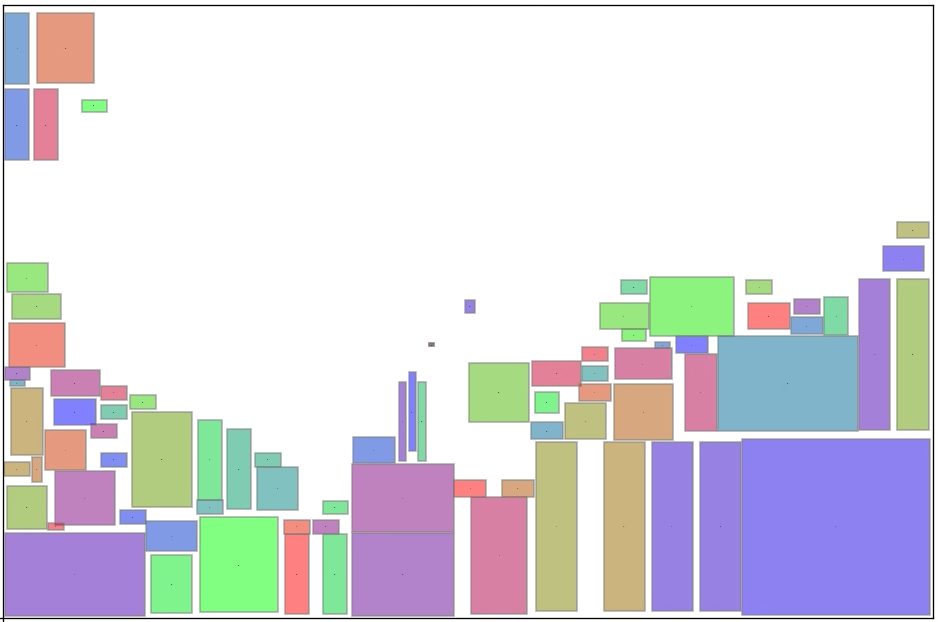}
         \caption{MOPPO \\
         $\omega=(1,0)$.}
     \end{subfigure}\hfill
     \begin{subfigure}[b]{0.12\textwidth}
         \centering
         \includegraphics[width=\textwidth]{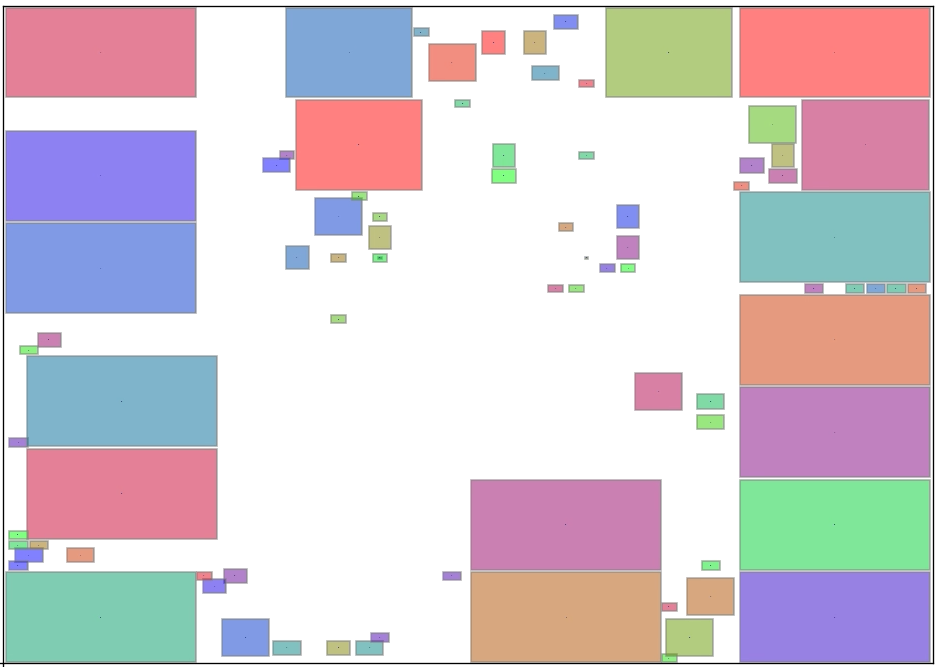}
         \includegraphics[width=\textwidth]{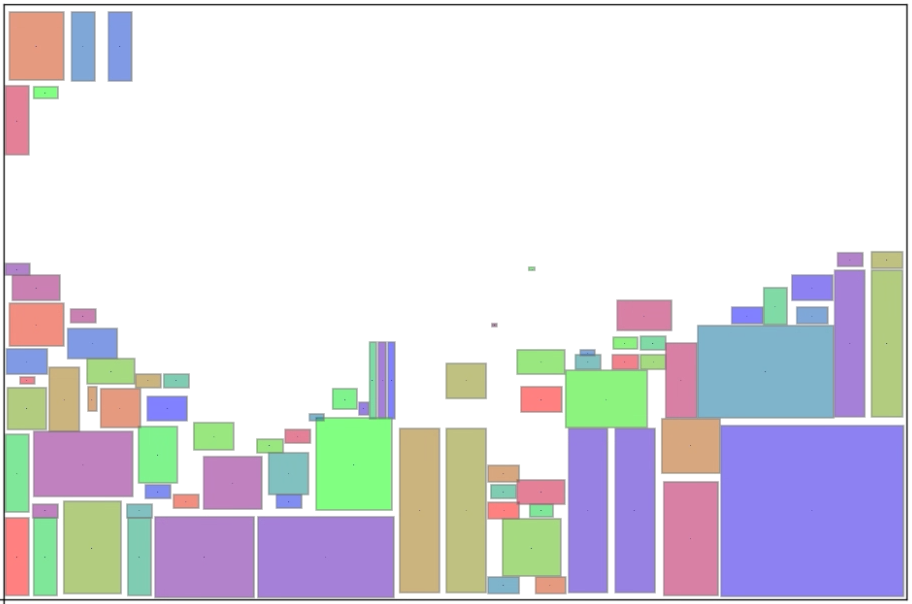}
         \caption{fixed preference\\
         $\omega =(1,0)$.}
     \end{subfigure}\hfill
    \begin{subfigure}[b]{0.12\textwidth}
         \centering
         \includegraphics[width=\textwidth]{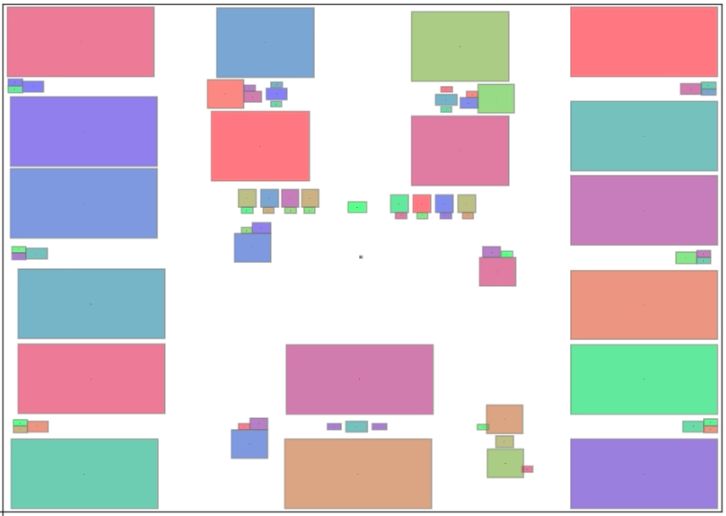}
         \includegraphics[width=\textwidth]{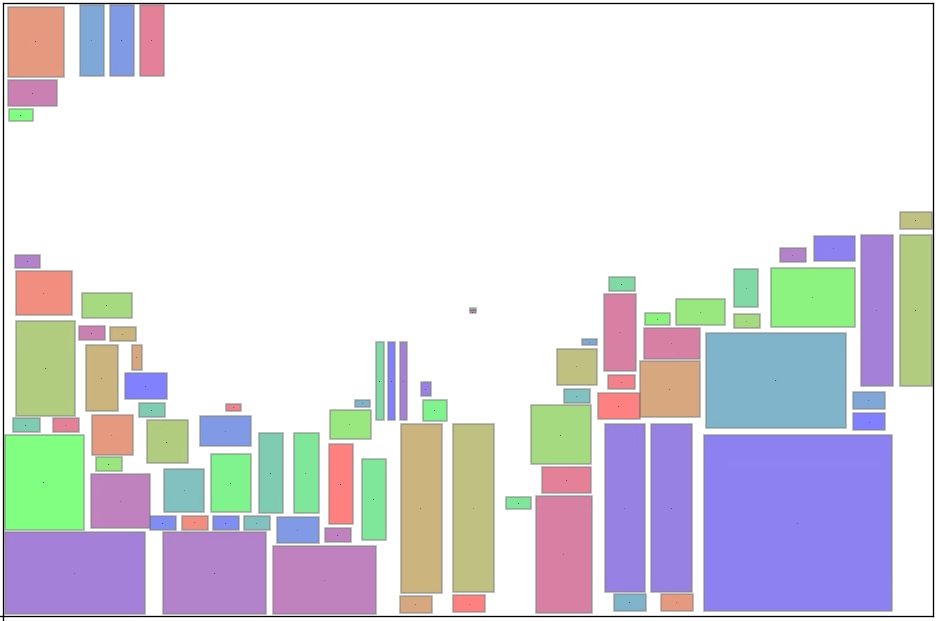}
         \caption{tapeout \\
         $\omega = N/A$.}
     \end{subfigure}
\caption{Placements of MOPPO, fixed-preference training and tapeout. First row: Superblue10; Second row: Superblue18.}
\label{fig:moppo_placement_result}
\end{figure*}

\begin{figure}[htbp]
\centering
\includegraphics[width=\columnwidth]{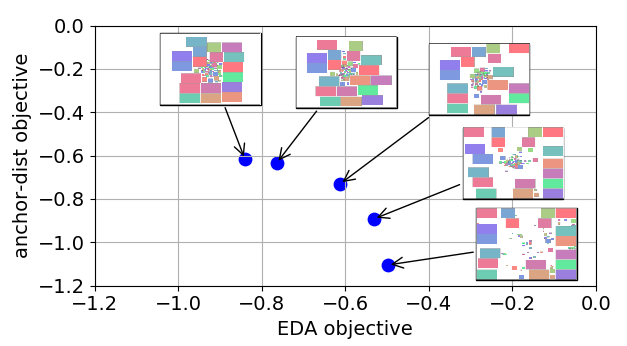}
\caption{Pareto frontier for a 1-dimensional objective simplex corresponding to our placement result.
}
\label{superblue-pareto}
\end{figure}

\subsection{Reward computation}\label{reward_comp}

To compute the reward efficiently, the netlist needs to be first clustered to speed up global placement and routing. We first use multilevel min-cut partitioning~\citep{DBLP:journals/tvlsi/KarypisAKS99} to create clustered netlist. We find that the number of clusters impacts the correlation of the computed reward before and after clustering. Therefore, we select the top 20 ranking macro placements from an industry placer~\citep{DBLP:conf/vlsi-dat/ChenLC20} that can output multiple candidate results for each circuit to analyze the most suitable cluster number for each circuit. Fig.~\ref{fig:cluster_size} shows wirelength and congestion correlation when using different cluster sizes on the testcase Superblue10. Since we want to keep a high correlation ($> 0.85$), we choose a cluster size of 3000 for this circuit. For other test cases, we also use the above methodology to decide the cluster size to find the minimum cluster size while keeping a good correlation for reward computation. The cluster size will affect the reward computation runtime. Typically we choose a size less than 5000 clusters, which allows rewards to be computed in under 10 seconds. 


After clustering, an analytical wirelength-driven placement~\citep{DBLP:journals/tcad/HuangLLYCCCCB18,DBLP:journals/tcad/LinJGLDRKP21} is applied to evenly distribute clusters. Then, a fast negotiation-based global router~\citep{DBLP:journals/tvlsi/DaiLL12} is used to obtain the congestion map. Since the design is clustered and g-cell size is adjusted according to the average cluster dimension, one-time reward computation can be done within 10 seconds. The rewards of wirelength $\ell_{WL}$ and congestion $\ell_{C}$ are computed according to Eq.~\ref{equ:wirelength} and Eq.~\ref{equ:congestion} in Section~\ref{metrics}. 
\subsection{Results}
In this section we present the results from the pretraining and zero-shot evaluation on Superblue10 and Superblue18.
Figure \ref{fig:moppo_placement_result} presents the resulting placements across different preferences. In Figs. 3(a) and 3(g) we observe the result of training standard PPO against a fixed preference. Fig. 3(h) presents the tapeout placement for comparison. This proves that our placements are competitive to human-designed placements and PPO fixed preference training on $(1,0)$ and $(0,1)$. The corresponding values are shown in Table~\ref{tab:results}.
Figure \ref{superblue-pareto} provides a visual representation of the Pareto frontier when sweeping through the preference $\omega$. The figures further exemplify the workflow of the design process facilitated by our tool, through which engineers may quickly search the preference space by evaluating placements at inference time, without the need for retraining.

\begin{figure*}[h]
     \centering
     \begin{subfigure}[b]{0.33\textwidth}
        \centering
\includegraphics[width=\columnwidth]{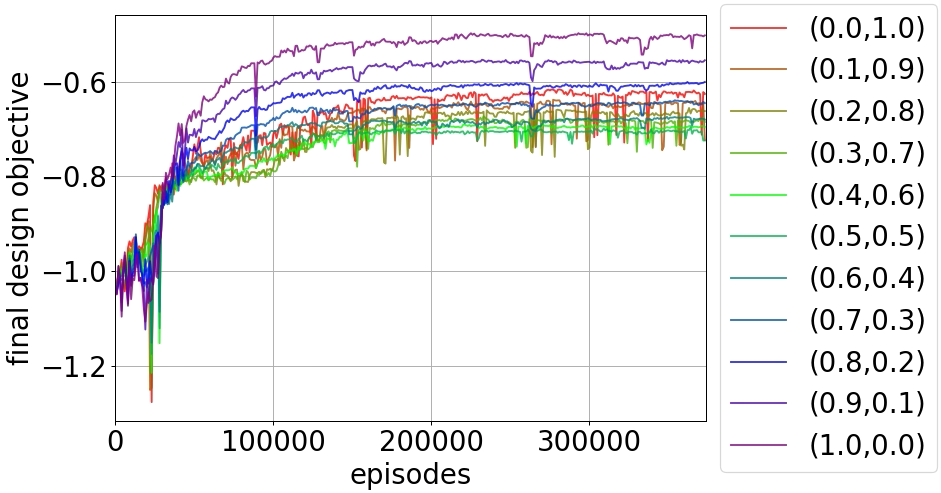}
         \caption{EDA objective}
         \label{fig:pref_final_design}
     \end{subfigure}\hfill
     \begin{subfigure}[b]{0.33\textwidth}
        \centering
\includegraphics[width=\columnwidth]{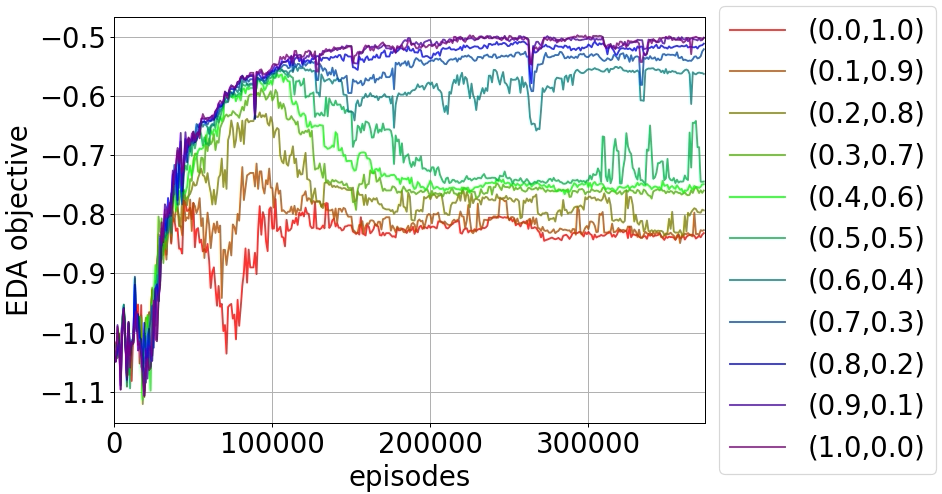}
         \caption{$\ell_{WL+C}$}
     \end{subfigure}\hfill
     \begin{subfigure}[b]{0.33\textwidth}
         \centering
         \includegraphics[width=\textwidth]{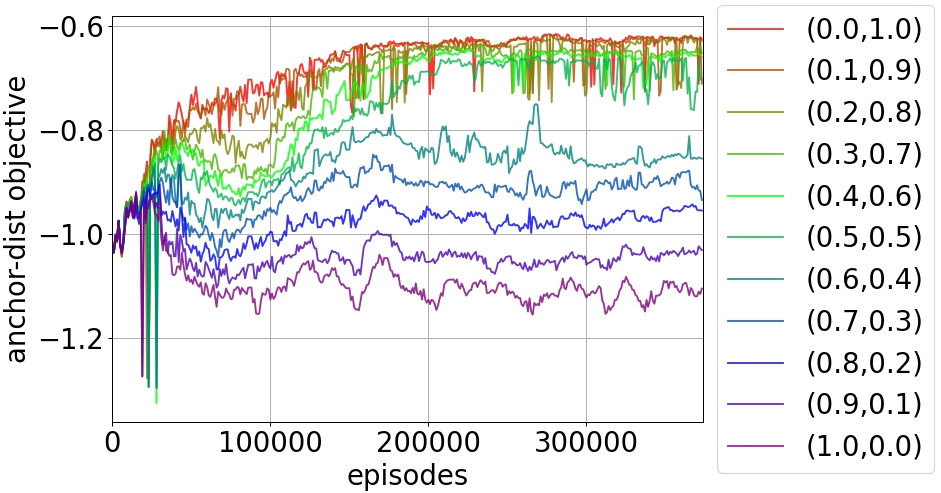}
         \caption{$\ell_{A}$}
     \end{subfigure}
\caption{ 
Plots of objectives during MOPPO training, evaluated by zero-shot inference, on preferences ranging from $(0,1)$ to $(1,0)$ on Superblue10.
}
\label{fig:zero_shot_eval}
\end{figure*}

Figure \ref{fig:zero_shot_eval} presents the zero-shot inference evaluation of the model during pretraining (at various model checkpoints), on a selected set of preferences. We observe the joint optimization of the objectives is possible for a period during training, after which the algorithm begins to discriminate between the two objectives and optimizes them selectively based on the preference. In Fig. 5(a) the combined EDA objective Eq.~\ref{EDA-obj} is evaluated for multiple preferences. In Fig. 5(b), the wirelength \& congestion objective is evaluated. Note that the preference $\omega = (1,0)$ corresponding to this objective alone, produces training curves that are on average decreasing. The opposite preference, $\omega = (0,1)$ sees progressively worse evaluation after a point. In panel Fig 5.(c), evaluating the anchor-distance objective, the situation is reversed.   

\begin{figure}[h]
     \centering
     \begin{subfigure}[b]{0.235\textwidth}
        \centering
\includegraphics[width=\columnwidth]{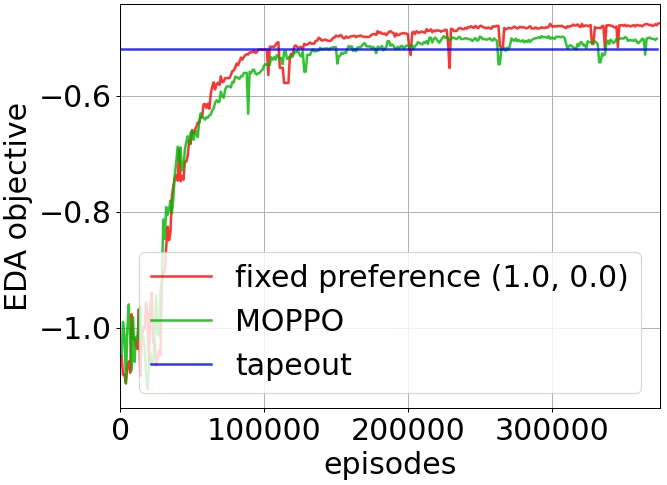}
         \caption{$\ell_{WL+C}$}
     \end{subfigure}\hfill
     \begin{subfigure}[b]{0.235\textwidth}
         \centering
         \includegraphics[width=\textwidth]{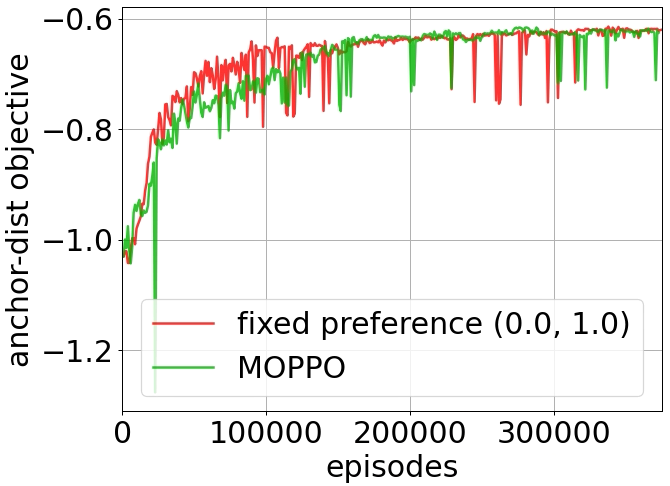}
         \caption{$\ell_{A}$}
         \label{fig:toy_reward_2}
     \end{subfigure}\hfill
\caption{Speed of convergence of MOPPO training compared to fixed-preference training on Superblue10 design. In panel (a), the value of the tapeout is presented as a threshold.}
\label{fig:speed_of_convergence}
\end{figure}

In Figure \ref{fig:speed_of_convergence} we present the relative speed of training MOPPO versus fixed-preference training. In both panels, though MOPPO converges more slowly than the fixed-preference training, it remains on the same order of magnitude, while capturing information about all preferences simultaneously. In Fig. 6(a), $\ell_{WL+C}$ can be competitively optimized to similar values as fixed-preference PPO.
In Fig. 6(b), the corresponding evaluation for $\ell_{A}$ is shown. The MOPPO curve comes from the same training run, and no adaptation of parameters is performed, whereas the fixed-preference training curve is rerun to optimize anchor-distance only. Overall we see that MOPPO makes only small sacrifices in its ability to optimize individual objectives versus the individually optimized PPO, which agrees with the definition of approximate Pareto optimality.

\section{Conclusions}
\label{conclusions}
In this paper, we presented a multiple-objective reinforcement learning solution for objective functions with inference-time variable weights using a single model, MOPPO. To speed up MOPPO training, we used trial macro placements to decide a suitable cluster size for fast reward computation and keeping high correlations to the rewards of unclustered netlists. We tested our method on complex mixed-size placement problems, where we have shown that our method can recover the Pareto frontier of two uncorrelated objectives in a single round of training. 
Furthermore, our method surpassed human placements in terms of $\ell_{WL+C}$, and achieved $\epsilon$-approximate optimality, i.e., remained close to the PPO optimized for a single fixed preference at all times.
Our research was built upon already successful frameworks and served to facilitate RL research in EDA by empowering engineers with a tool that they can quickly adapt to design constraints. However, it was not without its limitations. The training time of MORL was still very high even for the relatively simple problem of estimating a 1D Pareto frontier. For higher dimensional objective simplices, more optimization is required. Furthermore, the problem of correctly parametrizing policy families to guarantee  $\epsilon$-approximate optimal contour finding is still unsolved. Nevertheless, to the best of our knowledge, this paper represents a first step to bringing these methods together both in the context of on-policy RL as well as EDA in general.

\begin{table*}[htbp]
\footnotesize
  \centering
  \caption{Values of MOPPO placement objectives, compared to fixed-preference PPO and the tapeout. Similar values for wirelength and congestion between comparable placements of MOPPO, PPO and the tapeout are highlighted.}
        \begin{tabular}{ |c | c | c | c | c | c | c |c|c|c|c|c|}
        \hline 
        \multicolumn{2}{|c|}{\textbf{Design}}&
        \multicolumn{5}{c|}{Superblue10}&
        \multicolumn{5}{c|}{Superblue18}
         \\\hline
       \textbf{algorithm} &
        \textbf{$\omega$} & \shortstack{\textbf{final design} \\ \textbf{objective}}& \shortstack{\textbf{EDA} \\ \textbf{objective}} &\shortstack{\textbf{wirelength} \\ \textbf{$({\mu}m)$}}& \shortstack{\textbf{congestion}} & \shortstack{\textbf{anchor-dist} \\ \textbf{objective}}  &
        \shortstack{\textbf{final design} \\ \textbf{objective}}& \shortstack{\textbf{EDA} \\ \textbf{objective}} &\shortstack{\textbf{wirelength} \\ \textbf{$({\mu}m)$}}& \shortstack{\textbf{congestion}} & \shortstack{\textbf{anchor-dist} \\ \textbf{objective}} \\  
        \hline
        fixed preference   & $(0,1)$    & -0.61904  & -1.09231  &  535,984,538 &  1.48320  & -0.61904  &-0.55374  & -0.98822 & 112,955,805 &  1.73358 & -0.55374 \\ \hline  
         MOPPO              &  $(0,1)$   & -0.61609  & -0.83989  &  411,299,462 &  1.36695  & -0.61609  &-0.55663  & -0.91286 & 104,319,121 &  1.63929 & -0.55663\\ \hline  
         MOPPO              &  $(.3,.7)$ & -0.67289  & -0.76374  &  374,066,032 &  1.22565  & -0.63395  &-0.66294  & -0.80394 & 91,949,245 &  1.31816 & -0.60251\\ \hline  
         MOPPO              &  $(.5,.5)$ & -0.69609  & -0.61181  &  324,103,944 &  1.07478  & -0.73036  &-0.69952  & -0.72577 & 82,985,480 &  1.22835 & -0.67927\\ \hline  
         MOPPO              &  $(.7,.3)$ & -0.63890  & -0.53058  &  259,264,902 &  1.03320  & -0.89165  &-0.68760  & -0.63788 & 72,895,996 &  1.14353 & -0.80362\\ \hline  
         MOPPO              &  $(1,0)$   & -0.49719  & -0.49719  &  \textbf{242,484,931} &  \textbf{1.10682}  & -1.10419  &-0.56264  & -0.56264 & \textbf{64,240,346} & \textbf{ 1.10270} & -1.11108 \\ \hline  
         fixed preference  &  $(1,0)$   & -0.47467  & -0.47467  &  \textbf{232,916,882} &  \textbf{1.08465}  & -1.17216  &-0.55225  & -0.55225 & \textbf{63,061,503} &  \textbf{1.06913} & -1.13321\\ \hline  
         tapeout           &  N/A   & -0.52025  & -0.52025  &  \textbf{253,945,277} &  \textbf{1.09447}  & -1.11121  & -0.59883 &  -0.59883  &  \textbf{68,294,105} & \textbf{1.30058} & -1.08997 \\ \hline  
        
        \end{tabular} \\
    \label{tab:results}%
\end{table*}




\small

\bibliographystyle{IEEEtran}

\bibliography{IEEEabrv,bibliography}

\clearpage

\end{document}